# Force-Based Robotic Imitation Learning: A Two-Phase Approach for Construction Assembly Tasks

Hengxu You[1], Yang Ye, Ph.D.[2], Tianyu Zhou, Ph.D.[3], Eric Jing Du, Ph.D.[4*]

[1] Ph.D. Student, Engineering School of Sustainable Infrastructure & Environment, University of Florida, 1949 Stadium Road 454A Weil Hall, Gainesville, FL 32611; email: you.h@ufl.edu
[2] Assistant Professor, Department of Civil and Environmental Engineering, Northeastern University, 457 Snell Engineering Center, Boston, MA 02115; email: y.ye@northeastern.edu
[3] Post-doctoral Researcher, Engineering School of Sustainable Infrastructure & Environment, University of Florida, 1949 Stadium Road 454A Weil Hall, Gainesville, FL 32611; email: zhoutianyu@ufl.edu
[4] Professor, Engineering School of Sustainable Infrastructure & Environment, University of Florida, 1949 Stadium Road 460F Weil Hall, Gainesville, FL 32611; email: eric.du@essie.ufl.edu; Corresponding Author.

## ABSTRACT

The drive for increased efficiency, safety, and cost-reductions in the construction industry has intensified interest in robotics and automation. While robots can mitigate labor-intensive tasks with basic programming, tasks demanding intricate manipulation, such as welding and pipe insertion, present considerable challenges. These tasks require refined adaptive force control, making them intricate to learn and often leading to slow or non-convergence in robotic training. To navigate these challenges, potential solutions include leveraging deep imitation learning, i.e., revised reinforcement learning from human demonstrations. However, conventional imitation learning methods, predominantly reliant on visual or positional cues, often fall short in construction assembly tasks such like pipe insertion, where visual inputs can be obstructed, and force information contains more features needed for the completion. Addressing this, the paper introduces a two-phase system framework that integrates human-derived force feedback into robot learning. The first phase encompasses real-time data capture from human operators using a robot arm synchronized with a virtual physical simulator simulation, facilitated by ROS-Sharp as the data-bridge module. In the second phase, the force feedback is translated into robot motion instructions. A generative method is then employed to embed force feedback within the reward function, steering the learning trajectory. The effectiveness of the proposed method in terms of robotic motor learning was gauged through metrics such as task completion time steps and success rates. The paper's contribution is a two-phase framework that provides a similar-to-real force-based collision simulation, allowing human experts to experience real-time force interactions. This enables the capture of realistic force-responsive demonstrations, ensuring that the collected data effectively guides robot learning for adaptive and precise manipulation in construction assembly tasks.

**Keywords:** Imitation learning, force-based controls, construction assembly.

# 1. INTRODUCTION

The construction industry, in its pursuit of enhanced efficiency, reduced costs, and increased safety, is increasingly seeking automation and robotic assistance [1-3]. Robots have shown enormous potential to alleviate repetitive, and dangerous tasks from human workers, such as assembly, infrastructure inspection, material handling and heavy rigging [4-6]. Integrating the artificial intelligence (AI) agent with a physical robotic system could further improve the precision, reliability, and consistency of operations with competent training [7, 8]. While AI-enabled robots excel in performing repetitive and predefined tasks, dexterous and complex tasks still pose a significant difficulty such as welding and pipe insertion [9, 10]. Training a robot to perform these dexterous tasks demands delicate manipulation and adaptive force control, which induces diversity and several potential actions leading to a substantial increase in the complexity of the learning process and resulting in slow convergence or lack of convergence [11]

To tackle the challenges of learning in high-dimensional action spaces, Imitation Learning (IL) based methods are applied to leverage demonstrations from human experts or proficient use of human demonstrations as a form of instruction and reduce the size of action spaces that need to be explored [12-14]. Generative Adversarial Imitation Learning (GAIL)[15] could further address some key limitations of traditional IL by mitigating distributional shifts, thus enabling better exploration and performance in unseen states and generalizing better to new tasks [15]. The current formats of these instructions are largely reliant on visual or positional cues [16, 17]. In a task like a pipe installation, the vision sight is often obscured, and positional information might not be sufficient due to the irregular and rough interior surface of the pipe, which makes it evident that the conventional forms of instruction are inadequate for such tasks.

The limitations inherent in capturing vision-based demonstrations underscore the need for a more sophisticated form of human demonstration in robotic learning. In this context, we argue that force-based feedback represents a more effective modality for imitation learning, particularly in tasks requiring fine motor control. Human operators heavily depend on force feedback for tasks that necessitate dexterous manipulation, highlighting its significance [18, 19]. Integrating force feedback into the learning process of the robot could be instrumental in enhancing its ability to perform complex tasks [20, 21]. Besides, despite imitation learning's ability to reduce early-stage errors by following expert demonstrations, the robot agent may still deviate from optimal behavior due to data noise, algorithmic constraints, or unforeseen environmental variations [22, 23]. To ensure that the training agents can be effectively instructed, the demonstrations are required to be high-quality and generalizable [24]. Ensuring that human operators experience realistic force interactions allows the collected demonstration data to be more representative of real-world conditions, making it more effective in guiding robotic. Without an environment that accurately simulates real-world force interactions, human demonstrations may not fully capture the nuanced responses required for effective robot learning. Thus, the primary challenge lies in how to provide realistic force-based responses during human data collection so that human expertise can naturally react to collisions and adjust actions based on force feedback.

To address these challenges, this paper introduces a groundbreaking two-phase system framework that capitalizes on human-derived force feedback, integrating this data into the robot's learning protocol. The first phase involves real-time data collection from human operators using a high-precision robotic arm equipped with force sensors, which is synchronized with a virtual robot in a Unity simulation environment. This setup allows for the detailed capture of force-feedback data during task execution, facilitated by the data-bridge module ROS# for seamless data transmission between the physical robot and the simulation. The second phase concentrates on learning with force-based guidance, where the force feedback data is processed and transformed into motion instructions for the robot, ensuring accurate feature extraction. The use of Generative Adversarial Imitation Learning (GAIL) to integrate this force data into the reward function is a novel approach that shapes the learning trajectory. The effectiveness of the proposed force-based imitation learning approach is validated through a range of performance metrics, including task completion time steps and success rates. The discussion also covers the benefits and potential limitations of applying this methodology in real-world scenarios, emphasizing its applicability and scalability.

## 2. LITERATURE REVIEW

### *2.1. Deep Learning in Robot Training*

Over the past decade, Deep Learning (DL) has emerged as a compelling approach for training robotic systems to accomplish a variety of complex tasks, such as navigation in unseen environments [25], manipulation with high-cognitive complexity tasks [26, 27] or high-dexterity tasks [28] and humanoid locomotion [29]. In the specific field of intelligent robot learning, the methods could be classified into two major groups: Deep Reinforcement learning (DRL) and Deep Imitation Learning (DIL) [30].

The basic idea of DRL-related algorithms is to induce the deep neuron network (DNN) as the decision kernel of the intelligent robot and interpolate a high-dimensional optimal projection as the policy to guide the robot to make actions based on current observations to achieve the pre-defined goals [23, 31]. During the training process, the agent simultaneously undergoes exploration, where it tests new actions to gather information, and exploitation, where it leverages existing knowledge for optimal decision-making. These intertwined processes, comprising the exploration-exploitation trade-off, are intrinsic to the agent's learning throughout the reinforcement learning process [32]. Previous studies have already shown the effectiveness of the exploration-exploitation trade-off in reinforcement learning in scenarios involving tasks with limited action spaces, such as obstacle avoidance [33, 34], navigation [35, 36] and object operation [37, 38]. With the appropriate design of rewards, observations and actions, the robot trained with DRL can handle various tasks under different scenarios, such as under-water robot operation [39], emergency structural inspection [40] and rescue searching [41]. However, when it comes to dexterous operations, the inherent complexities give rise to an expansive action space and the vastness of possible actions leads the DRL models to invest a significant portion of exploration time. The process can be inefficient as the model sifts through numerous sub-optimal actions before finding an effective strategy [42].

To reduce the difficulty of exploring in large action space, the DIL-based methods were proposed that could leverage the human or expert demonstration for robot agent training [25], including behavioral cloning (BC), inverse reinforcement learning (IRL) and generative adversarial imitation learning (GAIL) [43]. The agent is forced to mimic expert behaviors by observing the demonstration so that it can rapidly find the correct solution to complete the task [26]. The versatility of DIL methods has been recognized, finding applications in various robot-learning domains, such as skill transfer [44], human-robot collaboration [45], etc. Liang et al. propose the learning-from-demonstration method to conduct quasi-repetitive construction tasks with demonstrations from humans [46]. The method also verifies the effectiveness of demonstration data generated in the virtual environment. Since the DIL methods have the overfitting problem that the agent would totally learn from the limited knowledge in a single demonstration, the combined framework is proposed that uses DIL to first initialize the training and reduce the action space then apply DRL to further explore within the limited scope to ensure generalizability [47]. [48-51] have theoretically shown the improvement of combining DIL and DRL compared with single DRL or DIL methods. From the perspective of real-word application, Huang et al. develop a novel training framework to teach the robot to learn long-horizon tasks with a more efficient exploration strategy. The results show that the combination of DIL-DRL training could effectively increase the robot's learning speed with high mission completion rate [52]. The DIL-DRL methods leverage the benefits of small action space and adversarial exploration proficiency which could largely increase the training efficiency.

*2.2. Imitation Learning for Construction Robots*

Construction robots have significantly advanced over recent years, enhancing efficiency, safety, and precision on construction sites. These robots are designed to handle various tasks, from material handling to intricate assembly processes. The development of these systems involves integrating advanced technologies such as autonomous navigation, scene understanding, and real-time monitoring. For instance, in [53], the importance of advanced scene understanding and adaptive manipulation in construction robots is discussed. Techniques like Clustering and Iterative Closest Point (CICP) and Generalized Resolution Correlative Scan Matching (GRCSM) enhance the robot's ability to accurately perceive and model construction components, enabling precise manipulation and adaptation to dynamic environments. Similarly, [54] highlights the integration of GRCSM for accurate geometry modeling and adaptive work plan formulation, allowing robots to handle tasks such as joint filling with high precision. Another critical aspect of construction robotics is ensuring effective human-robot collaboration. [55] introduces a digital twin system that enhances collaboration by synchronizing real-time data between the physical robot and its virtual model. This system facilitates real-time monitoring and control, improving efficiency and safety. Furthermore, [56, 57] offers a significant advancement in human-robot collaboration by enabling real-time data exchange and synchronization. Path planning is another essential component, as demonstrated in [58]. This paper proposes a method that integrates predicted movements of construction workers to achieve safe and efficient collaboration. [59] focuses on improving safety and collaboration by accurately estimating robot poses in real-time using a deep convolutional network. Additionally, [60] showcases the development of a robotic assembly system designed to improve the safety and efficiency of steel beam assembly processes. This system utilizes innovative methods for rotation, alignment, bolting, and unloading, significantly enhancing the assembly process. While the innovations in construction robotics have significantly improved various aspects of construction tasks, the integration of imitation learning (IL) offers a unique and powerful approach to further enhance these capabilities. By enabling robots to learn directly from human demonstrations, IL bridges the gap between human expertise and robotic execution, facilitating the handling of complex and dynamic construction environments with greater precision and adaptability.

Imitation learning (IL) has emerged as a vital approach in enhancing the capabilities of construction robots. By leveraging human expertise, IL enables robots to learn tasks through demonstration, which is particularly useful in the complex and dynamic environments of construction sites. This method focuses on achieving high accuracy, efficiency, and adaptability in various construction-related tasks. The framework introduced in [14] exemplifies this approach. It leverages a digital training environment and a federated construction skill cloud database to transfer construction skills from human workers to robots. This system enables robots to perform tasks such as ceiling installation with high precision by learning from human demonstrations in a virtual reality setting. Another significant contribution is found in [61], which discusses a method for teaching robots to perform tasks like ceiling tile installation by representing trajectories using generalized cylinders with orientation. This approach allows the robot to adapt its actions based on the geometric and orientation data of human demonstrations, enhancing its ability to handle complex tasks. [62, 63] explore the use of visual demonstrations and reinforcement learning to train robots for tasks such as ceiling tile installation. The context translation model and reinforcement learning components enable robots to learn and execute construction tasks by observing human demonstrations, improving their performance and adaptability. [46] highlights the feasibility and effectiveness of using IL to train robots for construction tasks. By combining context translation with reinforcement learning, this approach allows robots to learn from human demonstrations and adapt to the varying conditions of construction sites.

The previous work has made advanced explorations on how to use imitation learning to train robots with visual demonstrations. The success of these approach shows the potential of using other types of demonstrations, such as tactile or kinesthetic feedback, to guide the learning process. By expanding the types of demonstrations used, construction robots can achieve even greater versatility and performance in a broader range of tasks. This could lead to further innovations in how robots interact with their environments and adapt to new challenges, ultimately pushing the boundaries of what is possible in the construction industry.

## 2.3. Demonstration for Imitation Learning

In the realm of robot training within imitation learning paradigms, demonstrations serve as foundational blueprints that guide the robot's learning process [13]. Demonstrations have been instrumental in bridging the gap between human intuition and robotic execution, effectively translating human expertise into actions that a robot can mimic [64]. Demonstrations not only reduce the need for extensive training but also provide a structured pathway for robots to learn complex tasks that might be challenging to learn from scratch using traditional reinforcement learning methods. The major demonstration types to be used in DRL could be roughly divided into visual demonstrations [65], force/tactile demonstrations [66] and other types such as kinesthetic demonstrations [67] or language demonstrations [68].

Visual demonstration primarily leverages images or videos to represent instructions on certain tasks [69]. Taking advantage of the detailed texture information captured from the environment, the visual data intuitively capture the spatial and contextual details of an environment, aiding in tasks that robotic tasks like object manipulation [70], navigation [71], and interaction [72]. [73] proposed a GAIL-based framework for long-horizon construction tasks, relying primarily on visual feedback to guide robots in object manipulation and placement in large-scale environments, such as installing window panels. While visual data have proven invaluable for a broad range of robotic tasks, this type of demonstration has inherent limitations when guiding dexterous operations, particularly those that involve fine-scale manipulations such as pipe insertion [74]. Predominantly, visual sensors capture macro-level information and might struggle to discern the nuanced movements and adjustments that are often crucial in delicate operations [75]. For tasks like pipe insertion, the tactile feedback, including subtle resistances or minute texture variations, provides vital cues that visual data might overlook since the changes could lead to small or even no detectable object shiftiness or deformation. Furthermore, minor surface irregularities, which can significantly influence the precision of dexterous tasks, can easily elude visual detection, especially in non-controlled environments such as lighting changes or camera view obscuring [76]. Moreover, the very nature of certain tasks can result in occlusions, where essential aspects of the operation become visually obscured [77]. While embedding a moveable visual sensor with the operation object, such as inside a pipe, could theoretically offer a vantage visual perspective, it introduces logistical hurdles, from potential interference with the operation to the difficulty in its subsequent removal. In summary, the visual demonstration is not qualified to provide instructive demonstration if the task is performed in a dynamically changing and high-interactable environment.

Force demonstration, on the other hand, offers a uniquely apt perspective for dexterous operations. Despite the loss of sufficient texture information, force feedback captures the tangible distinctions intrinsic to delicate tasks [78]. [79] utilized tactile force feedback, processed through an SVM, to classify different interaction states in human-robot interaction tasks, enhancing the robot's ability to respond based on categorized tactile inputs. For intricate maneuvers such as pipe insertion, the precise force interactions, provide critical information about alignment, orientation, and fit with gentle guidance or resistance. The tactile nature of force demonstration ensures that even in scenarios where visual cues are compromised or unavailable, the agent can still receive a clear and direct signal of its interactions with the environment [80]. Consequently, for dexterous tasks demanding high precision and sensitivity, force demonstration becomes an invaluable source of guidance, ensuring accuracy and adaptability.

## 3. SYSTEM DESIGN

### 3.1. Overall Architecture

This study develops a motion skill transfer system to infer human adjustment strategies during task execution, enhancing a robot's learning for similar tasks. To parameterize the knowledge from human experience and speed up the robot arm training process, we propose an integration workflow including two steps: human data collection and fused deep reinforcement learning as shown in **Fig.1**. The Franka Emika Panda robot is used for data collection.

In the data collection session, a human expert is requested to hold the robot arm and attempt to insert the inner pipe into the outer one. Initially, we configure the robot arm to employ gravity compensation, enabling the expert to manipulate the end effector with minimal effort. The force data exerted directly on the end effector by the human expert is captured in Cartesian coordinates as $F_{demo} = (f_x, f_y, f_z)$. This primary action data is crucial as it directly influences the interaction forces relevant to the pipe insertion task.

Simultaneously, the human expert, equipped with virtual reality (VR) goggles, observes a simulated pipe insertion scene within a Unity environment. Actions performed by the expert are mirrored by the virtual robot arm in the simulation, where interactions such as collisions are simulated, and feedback is provided to the real robot arm. This bi-directional data flow allows the expert to adjust actions based on both real-time physical and simulated feedback.

Both the expert's direct actions on the end effector and the environmental interactions observed within the simulation are recorded. These human data collection patterns inform the deep reinforcement learning phase, which utilizes the proximal policy optimization (PPO) strategy to train agents for tasks like insertion. The training phase is represented by red lines in **Fig.1**.

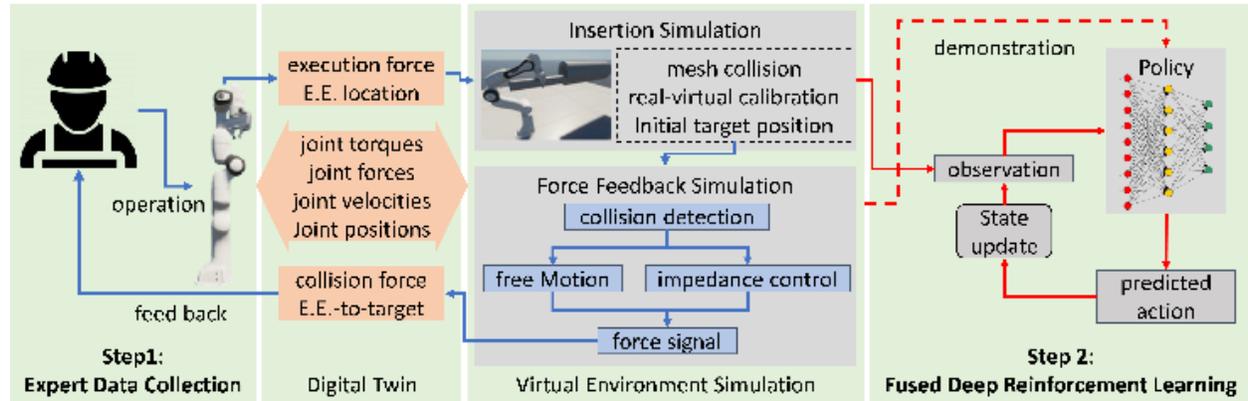

**Fig. 1**. The data pipeline of the three-step system.

*3.2. System Structure*

The system comprises two phases: Data Collection through Interaction (DCI) and Learning with Guidance (LG). In DCI, a human operates the FEP robot arm to perform tasks like pipe insertion, with actions and system states recorded for demonstration. The real arm and its Unity environment digital twin synchronize, transmitting virtual collision forces back as haptic feedback, aiding humans to refine their actions for quality data. In LG, these demonstrations guide a reinforcement learning agent in Unity, initially mimicking human actions, then evolving its policy through environmental rewards and penalties. This process, illustrated in **Fig.2**, shows DCI (black lines) and LG (orange lines) structures.

The system is divided into three functional layers: the interface, module, and algorithm layers. The interface layer establishes a real-time interactive environment for data collection and physical interaction in both DCI and LG phases. The module layer handles haptic feedback computation, data recording, and DIL simulation in these phases. Lastly, the algorithm layer extracts general features from observations and identifies aligned actions to guide the training process.

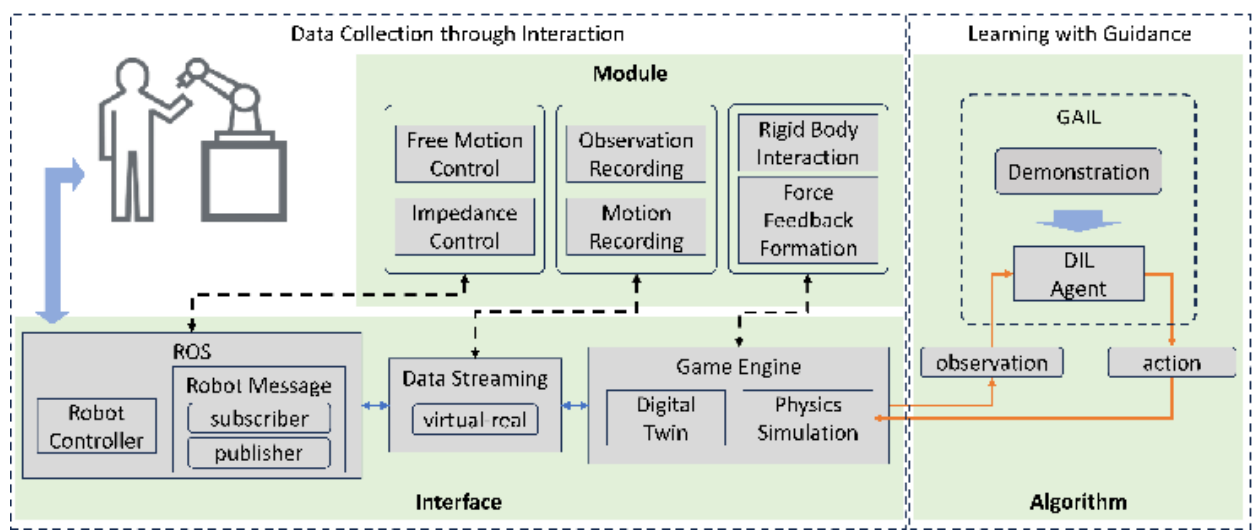

**Fig. 2**. The workflow of DCI and LG loops.

### 3.3. Robot Control for Collision and Non-Collision

In the DCI phase, the control module operates the real robot arm's background algorithm for human data collection. Experts control a virtual inner pipe using the robot arm's end effector in 3D cartesian space, with joint torques calculated via inverse kinematics (IK). This is visualized in **Fig.3**, which shows the panda robot arm's joint configuration. The real arm, primarily for data collection, is modified by removing the end effector and maintaining 6 Degree-of-Freedom (DoF), as depicted in **Fig.3(b)** with a vertical pipe attached to joint 6 as the user handle. Note that the handle used is a small, rigid metal pipe, chosen for its negligible deformation under operational loads. This ensures accurate force measurements and reliable manipulations. To integrate this handler with our robotic system, we designed a custom adapter to mount the pipe onto the Franka Emika Panda robot arm, replacing the original end effector. This secure connection minimizes mechanical delay and ensures precise operations. Details of the mounting mechanism and its impact on experimental integrity are illustrated in **Fig.3(c).**

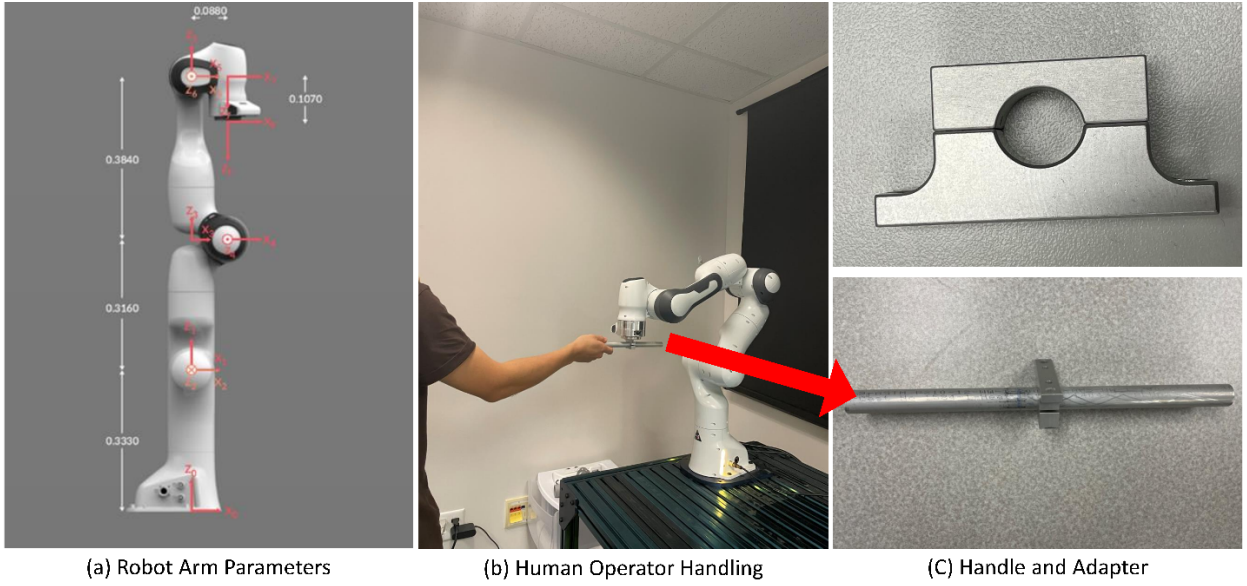

(a) Robot Arm Parameters    (b) Human Operator Handling    (C) Handle and Adapter

**Fig. 3**. (a) The physical configuration of the robot arm. (b) The setup for human operator. (c) Adapter details.

Two control modules, Motion and Force module, were developed to simulate the real operation condition corresponding to the to the collision and non-collision condition respectively. Let $x_0^t$ be the desired position of the handler at time stamp $t$ and $x_{dis}^t = x^{t^T} - x_0^{t^T}$ denotes the displacement between the handler's current and desired position at time stamp $t$, then the control algorithm could be expressed as:

$$\boldsymbol{F} = \delta_{mov}(\boldsymbol{x}^{t-1}, \boldsymbol{x}^t | \boldsymbol{x_0}^t) \cdot \delta_{coll}(\boldsymbol{x}) \cdot (M_c \cdot \ddot{\boldsymbol{x}}_{dis}^t + B_c \cdot \dot{\boldsymbol{x}}_{dis}^t + K_c \cdot \boldsymbol{x}_{dis}^t) + \boldsymbol{g}_c, \quad (1)$$

where $\boldsymbol{F}$ denotes the desired resistance force in cartesian space, $M_c$ denotes the inertia coefficient for robot's self-resistance, $B_c$ denotes the damping coefficient related to the speed of motion, $K_c$ denotes the damping coefficient for motion speed and $g_c$ refers to the constant force compensating for gravity. $\delta_{mov}(\cdot,\cdot|\boldsymbol{x_0}^t)$ and $\delta_{coll}(\cdot)$ adjust control modes based on simulation results. $\delta_{mov}(\boldsymbol{x}^{t-1}, \boldsymbol{x}^t | \boldsymbol{x_0}^t)$ predicts whether the inner pipe is moving towards or away from the last collision point at time stamp $t$, with $x_0^t$ being the desired position when collision happens and $x^{t-1}, x^t$ being the position of handler at time stamp $t-1$ and $t$, respectively. $\delta_{coll}(\cdot)$ indicates collision status, taking the handler's 3D position and outputting 0 for non-collision and 1 for collision. The detailed calculation and application of $\delta_{mov}(.,.|x_0^t)$ will be discussed in the next paragraph and $\delta_{coll}(\cdot)$ will be further discussed in the next section.

For the non-collision, we set $\delta_{coll}(x^t) = 0$, indicating that the inner pipe held by the virtual handler doesn't collide with the outer pipe at position $x^t$. Note that the virtual and real robot arms collaborated so $x^t$ is also the coordinate of the virtual handler. The expert can move the handler without any resistance and the robot will naturally calculate the desired joint positions according to **Eq (2)** to follow the expert's movement. In this circumstance, only the gravity compensation force is applied to maintain the robot's current position.

For the collision condition, we set $\delta_{coll}(.) = 1$. The desired position $x_0^t$ will be a temporal constant position $x_0$ which equal to the handler's position when collision happens. It will remain the same until the inner pipe and outer pipe are separated. By setting a very large $K_c$ in impedance control, we can simulate the static resistant force between two colliding rigid bodies. However, this approach causes a large resistant force in all directions of translation away from $x_0$. When the handler moves away from the collision point, there should be no resistant force. To address this, we introduce $\delta_{mov}$, which indicates whether the handler is moving away from the collision point. This allows us to determine whether or not the resistant force should be applied. At the time stamp t we calculate the indicator's value as:

$$\delta_{mov}(x^{t-1}, x^t | x_0) = u(\frac{(x^t - x^{t-1})^T}{||x^t - x^{t-1}||} \cdot \frac{(x^{t-1} - x_0)}{||x^{t-1} - x_0||})$$
$$u(x) = \begin{cases} 1, & if\ x \geq 0 \\ 0, & if\ x < 0 \end{cases} \quad (2)$$

Specifically, we assume that the collision has already occurred at $t-1$. The vector $(x^{t-1} - x_0)/||x^{t-1} - x_0||$ indicates the direction that could lead to further collision and $(x^t - x^{t-1})/||x^t - x^{t-1}||$ indicates the handler's movement trend at $t$. If the handler is still moving in a direction that could lead to further collision at time $t$, $\delta_{mov}$ is set to 1, allowing the large resistant force to remain active, simulating the contact force. Conversely, if the handler is moving away from the collision point, $\delta_{mov}$ is set to 0, meaning no resistant force is applied.

*3.4. Collision Simulation*

The pipe insertion task poses significant computational challenges for Unity's PhysX engine due to the continuous contact between the inner and outer pipes. While PhysX is effective for basic physical interactions, it struggles with prolonged collisions, leading to resource-intensive processes, delays, and inaccuracies like object penetration. These issues disrupt simulation realism and compromise the quality of demonstration data, rendering it unsuitable for training. To address this issue, we developed a simplified force simulation system tailored specifically for contact-rich tasks like pipe insertion. Instead of relying entirely on PhysX for resolving collisions, we leveraged PhysX's OnTrigger function, which allows for precise detection of collision points without invoking the computationally expensive physics solver. In addition, we employed impedance control to simulate the collision forces. Even when objects are colliding, slight translations occur if the objects are pushed together by a large force. These small movements are used to calculate the velocity of the objects, which, combined with their mass, allows us to infer force feedback through a custom mathematical model. This method efficiently replicates the physical dynamics of the task without the need for constant collision recalculations. The process involves collision detection and impact resolution, yielding realistic output force on colliding objects. Detailed explanations follow in subsequent sections.

*3.4.1 Collision Simulation*

During collision detection, the simulator identifies contact points between object colliders in two stages: broad-phase and narrow-phase detection. Broad-phase is a preliminary step identifying potential colliding objects. If a possible collision is detected, narrow-phase detection accurately simulates the collision. The simulator checks for intersections between mesh colliders; if none are found, $\delta_{coll}(x) = 0$. Otherwise, $\delta_{coll}(x) = 1$. The value will be transferred to the control module to control the real robot arm as mentioned before. Upon intersection detection, contact points and their normal vectors (pointing outwards from contact points) are calculated. **Fig.4** illustrates these points (red dots) and vectors (white lines). Contact points are denoted as $pt_i \in R^3$ with corresponding normal vector $\boldsymbol{n_i}$ and the results are represented as $PT = \{(pt_i, \boldsymbol{n_i}) | i \in 1, \ldots, N\}$ where $N$ is the number of total detected points.

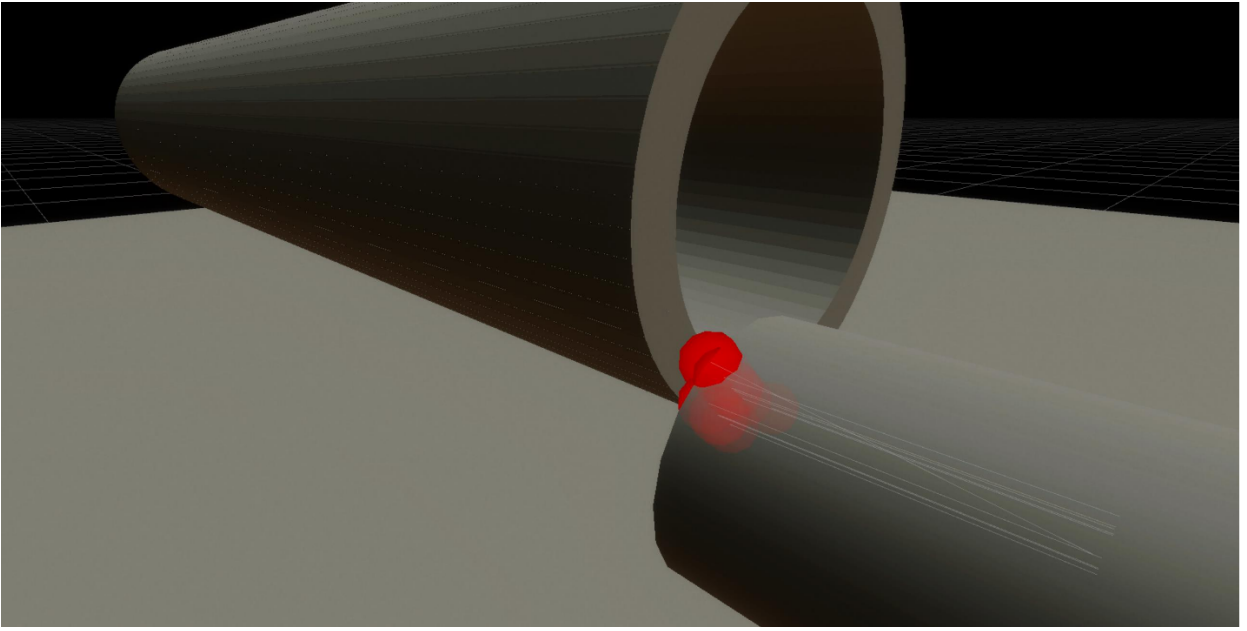

**Fig. 4**. The detected collision points of the two colliders.

### *3.4.2 Collision Resolution*

To support real-time interaction response, we applied a customized trigger method to resolve the collision interaction that only provides the detection results with simplified physical interaction rules. Specifically, we split the total collision process into two phases: pre-collision during frame $t-1$ to $t$ and on-collision during frame $t$ to $t+1$. For the pre-collision phase, we applied Newton's second law to calculate the impulse that only lasts for a single frame. Let $\boldsymbol{x}^{t-1}$ and $\boldsymbol{x}^t$ be the locations of the handler at time stamps t-1 and t, respectively. The moving speed of the handler is calculated by:where $T^{t-1,t}$ is the time difference between the two adjacent frames. Let $m$ be the pre-defined mass of the inner pipe in the simulation environment, then the impulse magnitude on the inner pipe at frame $t$ is calculated by:

$$F_{pre}^t = \frac{m \cdot |\boldsymbol{v}^t|}{T^{t-1,t}}. \tag{3}$$

To estimate the normal of the closest contact point, we attach a ray-caster on the inner pipe at the exact location of the closest contact point. The ray-caster will emit $n_{ray}$ rays that return $n_{ray}$ points on the outer pipe surface centered around $\boldsymbol{pt}^t_{closest}$ distributed in a spherical shape. To ensure that all the ray-caster points are on the outer pipe surface, the angles between rays and the vector $\overrightarrow{\boldsymbol{x}^t \cdot \boldsymbol{pt}^t_{closest}}$ is set to be smaller than 45 degrees.

Given the set of ray-caster points as $\{\boldsymbol{pt}^t_{ray_i} | i \in [1, \dots, n_{ray}]\}$, the center is calculated by:

$$\boldsymbol{pt}^t_{ray_{cen}} = \frac{\sum_{i=1}^{n_{ray}} \boldsymbol{pt}^t_{ray_i}}{n_{ray}}, \tag{4}$$

and the normalized point sets are calculated as:

$$\overline{\boldsymbol{pt}^t_{ray_i}} = \boldsymbol{pt}^t_{ray_i} - \boldsymbol{pt}^t_{ray_{cen}} \quad for\ i\ in\ range(n_{ray}). \tag{5}$$

The normalized points are used to construct the matrix $\boldsymbol{A}$ as:

$$\boldsymbol{A} = \begin{bmatrix} \boldsymbol{pt}^t_{ray_1}.x & \boldsymbol{pt}^t_{ray_1}.y & \boldsymbol{pt}^t_{ray_1}.z \\ \vdots & \vdots & \vdots \\ \boldsymbol{pt}^t_{ray_{n_{ray}}}.x & \boldsymbol{pt}^t_{ray_{n_{ray}}}.y & \boldsymbol{pt}^t_{ray_{n_{ray}}}.z \end{bmatrix},$$

and derived the singular value decomposition (SVD) results of $\boldsymbol{A}$ as:

$$\boldsymbol{A} = \boldsymbol{U} \cdot \boldsymbol{S} \cdot \boldsymbol{V}^T. \tag{6}$$

Then the surface normal is the last column of $\boldsymbol{V}$ which corresponds to the least singular vector of $\boldsymbol{A}$, denoted as $\overline{\boldsymbol{n}^t_{closest}}$, and the direction of the impulse is calculated by:

$$\overline{\boldsymbol{F}^t_{closest}} = \overline{\boldsymbol{n}^t_{closest}} - 2 \cdot (v^t * \overline{\boldsymbol{n}^t_{closest}}) * \overline{\boldsymbol{n}^t_{closest}}, \tag{7}$$

and the impulse force vector is calculated by:

$$\boldsymbol{F}^t_{pre} = F^t_{pre} \cdot \overline{\boldsymbol{F}^t_{closest}} \tag{8}$$

In the on-collision phase with constant force, both inner and outer pipes being rigid bodies, the contact force equals the external force on the inner pipe, calculated using **Eq (8)**. Here, not just the colliding force, but also friction, is crucial for guiding the expert. The variation in friction provides haptic feedback, indicating the inner pipe's position: increasing friction suggests an improper position. The normal supporting force and friction are derived using specific formulas as:

$$\begin{aligned} F_{normal}^t &= F_{on}^t * \overline{n_{closest}^t} \\ F_{friction}^t &= F_{on}^t - F_{normal}^t \end{aligned} \tag{9}$$

To ensure ease of separation between rigid bodies during expert manipulation, a shifting prediction module is implemented. This prevents experts from feeling unrealistic resistance when moving the inner pipe away from a contact point, as would be suggested by **Eq (8)**. To address this, a motion prediction indicator $\delta_{mov}(\cdot,\cdot)$ is used. If the collision happens at time stamp $t$ and the current location is denoted as $x_t$, then at the time stamp $x^{t+1}$ we calculate the indicator's value as:

$$Ind = (x^{t+1} - x_0) * (x^t - x_0). \tag{10}$$

If $Ind \geq 0$, $\delta_{mov}$ is set as 0. On the other hand, $\delta_{mov}$ is set as 1, indicating that the expert is still trying to push the inner pipe to the collision point.

### 3.4.2 Deep Learning Process

Similar to the traditional reinforcement learning problem, the operation sequence of the pipe insertion task is formed as a Markov decision process (MDP), denoted as $M = (S, A, P_a, R_a)$, where $S$ is the set of possible states collected from the current environment, $A$ is the set of potential actions that the agent could take, $P_a$ is the probability that the agent would take action $a$ with the current observation and $R_a$ is the expected immediate reward set. Two types of observations are used for comparison: the traditional visual data and the simulated force data proposed in this study which will be discussed in the following section. The specific optimization target for pipe insertion is then formed as inserting the inner pipe into the outer pipe with a certain depth with the minimum number of steps.

Then, during the data collection process, the human expert is required to hold the handler and try to insert the inner pipe into the outer pipe at a certain depth. The observations and collision feedback are recorded as a demonstration using the ML-Agent imitation learning package. Two types of demonstration data are used with the corresponding training setup for comparison, including visual demonstration and force demonstration.

For the force demonstration, the depth and distance information are both provided. The distance information used in this context refers to the distance between the inner pipe's tip (edge) and the outer pipe's center. The depth information refers to the moving distance along the insertion direction, which is critical for tracking the progression of the insertion and ensuring proper positioning throughout the task. Additionally, the force observation $\boldsymbol{F}_{ob} = [\boldsymbol{F}_{normal}{}^T, \boldsymbol{F}_{friction}{}^T]^T$ is included as calculated in **Eq (8)**. The human expert could also monitor the insertion process through a global camera outside as shown in **Fig.5(a)**.

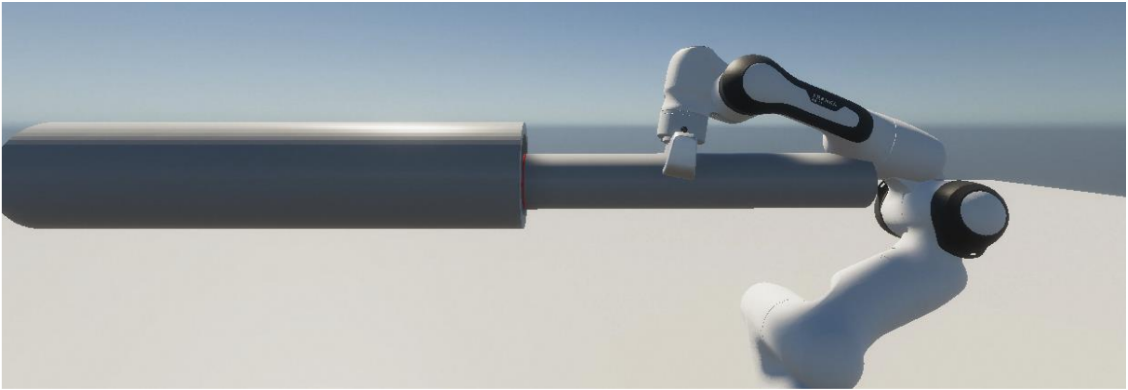

(a) Global viewing camera

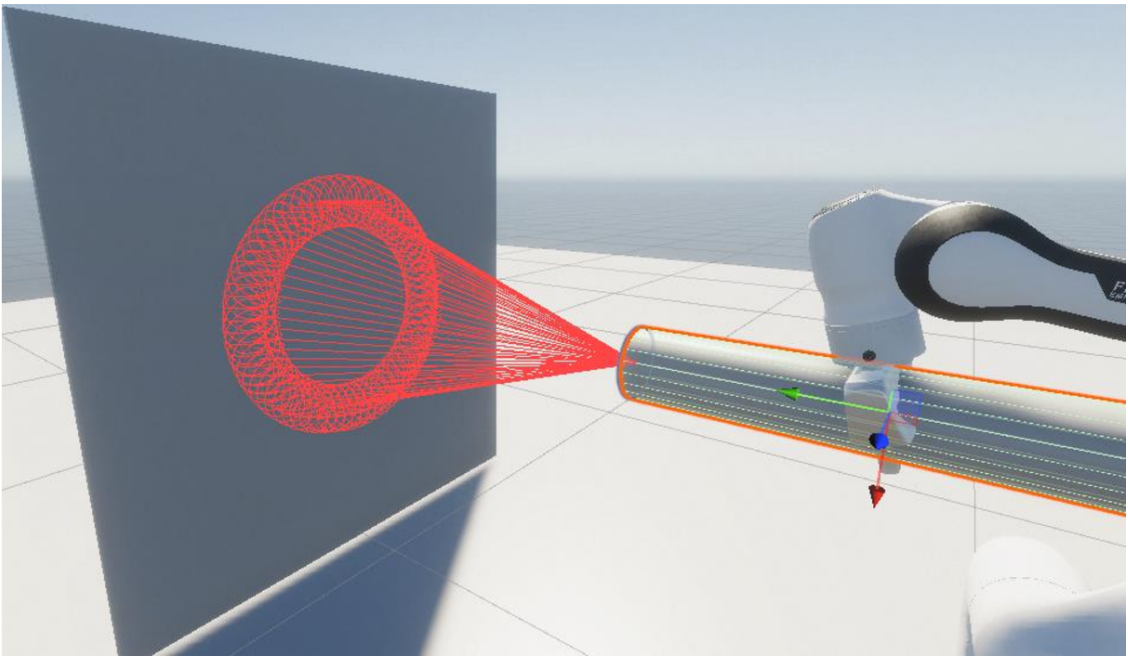

(b) Raycasters arrangement

**Fig. 5**. Illustration of data collection.

For the visual demonstration, a visual sensor is embedded near the closure of the inner pipe, facing the outer pipe to ensure continuous visual feedback during insertion. The visual feedback is sourced from 64 Raycasters, capturing both the positions of reflection points and the associated object types they intersect with. Specifically, the observation for visual demonstration is formed as $Vis_{ob} := \{d, l, [P]\}$, where $d$ is the insertion depth, $l$ is the distance from the inner pipe to the target and $P \in R^{64*4}$ are the reflection points with the distances and the interaction objects' types denoted by the integer index. **Fig.5(b)** illustrates the arrangement of Raycasters and the point clouds from the human expert's view. The global view is also provided during the visual-based insertion process.

The training environment, built using the ML-Agent package, involves two phases: deep imitation learning for initialization and DRL for fine-tuning and exploration. During initialization, Generative Adversarial Imitation Learning (GAIL) is employed. Following GAIL, the agent enters the DRL phase, using Proximal Policy Optimization (PPO), configured as per the ML-Agents toolkit, for behavior refinement and enhanced adaptability.

## 4. EXPEERIMENTS

The physical interaction simulation system is designed in Unity using the physical engine. In our experiments, the outer pipe exists only in Unity and its position is directly obtained, while the digital twin is built for the robot arm to collect human expert action data through VR. The training process is built with ML-Agents which is a well-designed reinforcement learning framework that could be directly plugged into the physical environment and collected data through the interaction. To assess the success of the pipe assembly task, the success criterion is set as the inner pipe being inserted into the outer pipe to a depth greater than 0.5m. A thin pad is placed inside the outer pipe at the 0.5m depth mark to detect collisions. If the inner pipe contacts the pad, it indicates that the task has been successfully completed.

The policy network to be trained is structured using the configuration file provided by ML-Agents (Juliani et al. 2018). For the GAIL model, the generator consists of an input layer, a single hidden layer and an output layer. The generator and discriminator network architecture employed in this work is adapted from established models commonly used in adversarial learning frameworks.

The input layer for the force demonstration group includes eight neurons: six for the observation vector $\boldsymbol{F}_{ob}$ and two for depth and distance. The visual demonstration group's generator has an input of 258 neurons, formed from a 4×64 dimensions vector plus two for depth and distance. Despite different input configurations, both groups use a common hidden layer with 256 neurons.

The output layers in both are identical, consisting of three neurons for Cartesian space force exertion. This design balances modality-specific inputs with a unified structure in hidden and output layers, capturing each feedback type's specifics while ensuring consistency.

The discriminator is formed by an input layer, two hidden layers and an output layer. The force group's discriminator has an 11-neuron input layer for the state-action pair $(s, a)$, with 8 for observations and 3 for actions. It features two 256-neuron hidden layers and a single-neuron output for confidence scoring. Similarly, the visual group's generator and discriminator mirror this structure. The visual group's discriminator has 261 input neurons with 258 for observations from 64 Raycasters (with each vector's first three elements one-hot encoding colliding object types and the last element recording collision distance) and 3 for actions. The final two neurons track depth measurements.

The deep reinforcement learning policy networks for force and visual groups are the same as the generators, correspondingly. The reward mechanism for the training is implemented as follows: Let $F_{normal_t}, F_{friction_t}$ denote the normal force and friction force at timestep $t$. Then the reward $r_t$ is defined as:

$$r_t = \begin{cases} \frac{1}{MaxStep}, & if \; ||F_{normal_t}|| < ||F_{normal_{t-1}}|| \; and \; ||F_{friction_t}|| < ||F_{friction_{t-1}}|| \\ 0, & if \; ||F_{normal_t}|| = ||F_{normal_{t-1}}|| \; and \; ||F_{friction_t}|| = ||F_{friction_{t-1}}|| \\ -\frac{1}{MaxStep}, & else \end{cases} \quad (11)$$

Additionally, if the pipe is inserted to the desired depth, a value of 1 score will be added to the total score of the current episode. For every step before reaching the target depth, a negative score with $-1/MaxStep$ will be added to the total score. The reward mechanism is designed with two main objectives: task completion and collision avoidance. The robot is rewarded with 1 upon successfully inserting the pipe to the desired depth as the task is completed. To prevent damage from collisions, the robot is given positive feedback when both the normal and friction forces decrease over time, indicating it is avoiding collisions. Conversely, if the robot moves toward the collision point, a negative score is assigned. The $1/MaxStep$ term is used for normalization, ensuring the reward remains within a reasonable scale and promotes stable learning.

For training, we chose a batch size of 128 and a buffer size of 2,048, with a learning rate of 0.0003 decreasing linearly. Inputs are normalized, and a simple visual encoding is used. The extrinsic reward has a strength of 1.0 and a discount factor of 0.99. Behavioral cloning occurs in the first 10,000 steps with a strength of 1.0, not updating regularly (samples per update set to zero). The GAIL reward signal is at 0.01 strength with a 0.99 discount factor. Training retains the last five checkpoints over a maximum of 5 million steps, with each episode lasting 10,000 steps and summaries generated every 12,000 steps.

## 5. LEARNING RESULTS

### 5.1. Overall Performance

We collected 20 human demonstrations for each group and trained the agent with each data sample respectively. **Fig.6** showed the reward plots of all 40 training trials with the green lines denoting the force demonstrations and the red lines denoting the visual demonstrations. The dotted black line is the reinforcement learning (RL) baseline without force or Raycaster visual observation, only the distance and depth information are provided. The solid purple line is the reward plot for force group baseline and the dashed purple line is for the visual group. The maximum cumulated reward for a single episode is approximately 1.4 and we set the total training step to be 10,000,000. The detailed training hyperparameters are listed in **Table 1.**

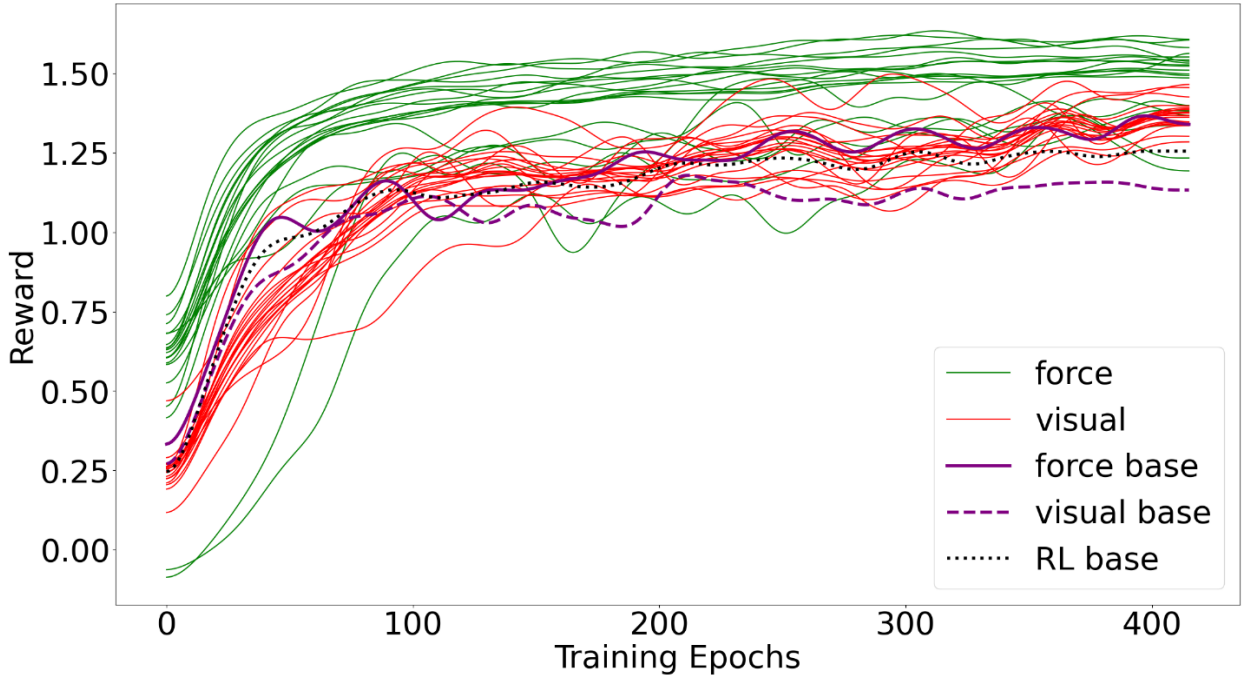

**Fig. 6.** Training results of all the demonstrations.

**Table 1.** Hyperparameters for Training

| Training Settings | | Network Setting | |
|---|---|---|---|
| trainer | PPO | hidden units | 128 |
| batch size | 128 | number of hidden | 2 |
| buffer size | 2048 | visual encoder | Conv_2d |
| number of epochs | 3 | **Update Setting** | |
| learning rate | 0.0003 | total steps | 5,000,000 |
| regularization | 0.95 | max steps | 10,000 |
| reward discount | 0.99 | summary steps | 12,000 |

To train the force and visual group samples, we employ the DIL-DRL training workflow, where the policy network is first initialized using deep imitation learning (DIL) and then further refined with deep reinforcement learning (DRL) for exploration and generalizability. The force baseline refers to the scenario where observations and actions are identical to those in the force group samples, but training is done directly with DRL, without DIL initialization. Similarly, the visual baseline is the case where observations and actions match those in the visual group samples, but training is conducted solely with DRL, bypassing the DIL initialization phase. According to the results in **Fig.6**, almost all the DIL-DRL training samples performed better than the baseline considering the convergent speed. For the green lines group (force samples) and red lines group (visual samples), it could be observed that the green lines, which are the force demonstrations, generally converge faster than the red lines, which are the visual demonstrations. The comparison of convergency speed fundamentally shows the benefits of using force demonstration to guide the learning process. Besides, the training results for both the RL and force-based conditions, as shown in **Fig.6**, quantitatively highlight the importance of incorporating perceivable force feedback. In the RL condition, the human expert does not receive force feedback during data collection, and the robot's observation during learning is limited to the handler's position relative to the robot arm's base coordinates. As a result, the reward at convergence is approximately 70% of that achieved in the force feedback condition. In the force-based condition, although collision force is included in the robot's observation, the robot is trained directly with DRL without first using the DIL phase to initialize the training with demonstration data collected from the perceivable force feedback system. The reward plot shows that all force-based examples (green lines) outperform the force-based condition trained solely with DRL (purple line), demonstrating the importance of both force feedback and imitation learning for improved performance.

To further evaluate the training performance, we illustrated the distributions of the two groups' training results in **Fig.7.** The red curves in the two plots show the means of the two groups and the shadow areas cover the 25%-75 performance segments. Similarly, the mean plot for the force group converges faster than that of the visual group. Additionally, Observing the 25%-75% performance segments, agents trained with force demonstrations consistently surpassed those trained with visual demonstrations. Trials utilizing force demonstrations yielded reward plots with narrower distributions, indicating more consistent agent performance. In contrast, those relying on visual demonstrations had wider reward distribution, pointing to greater variability in performance and potential challenges in generalization.

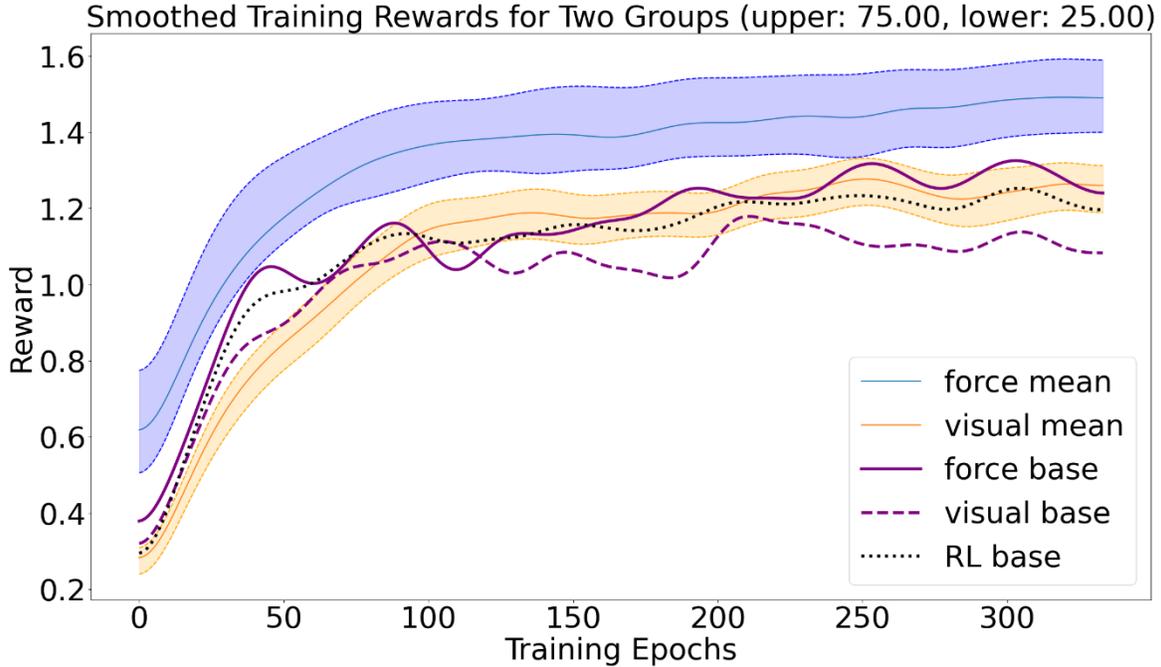

**Fig. 7.** Distribution of the training plots with inner pipe randomization.

*5.2. Condition 1: Randomized Inner Pipe Location*

To ensure the generalization of the learned policy, we randomly initialized the outer pipe and inner pipe positions during the inferencing process with the target position to be fixed. The distance between the target center and the initial position varies from 0.6m to 0.8m along the x-axis and falls within a circle with a 0.15m radius on the y-z plane. The solid purple line denotes the baseline with pure DRL for force observation and the dashed purple line for visual observation. The black dotted line is for the RL baseline.

We listed the inference results of the learned models from all demonstrations in each group and the baselines as shown in **Table 2** and **Table 3** for force and visual, respectively. The second row in each table listed the number of successes within 100 trials. the third row showed the average episode length for the successful trials and the fourth row showed the standard deviation (STD) error. For both the force baseline and visual baseline, the better performance is observed compared with the pure RL baseline, which indicates the importance of adding the force and visual observations. All the trials have a significantly higher accuracy compared with the baseline. For the two comparison groups, it is obvious that the force group performed better than the visual group on both the successful rate and the average episode length.

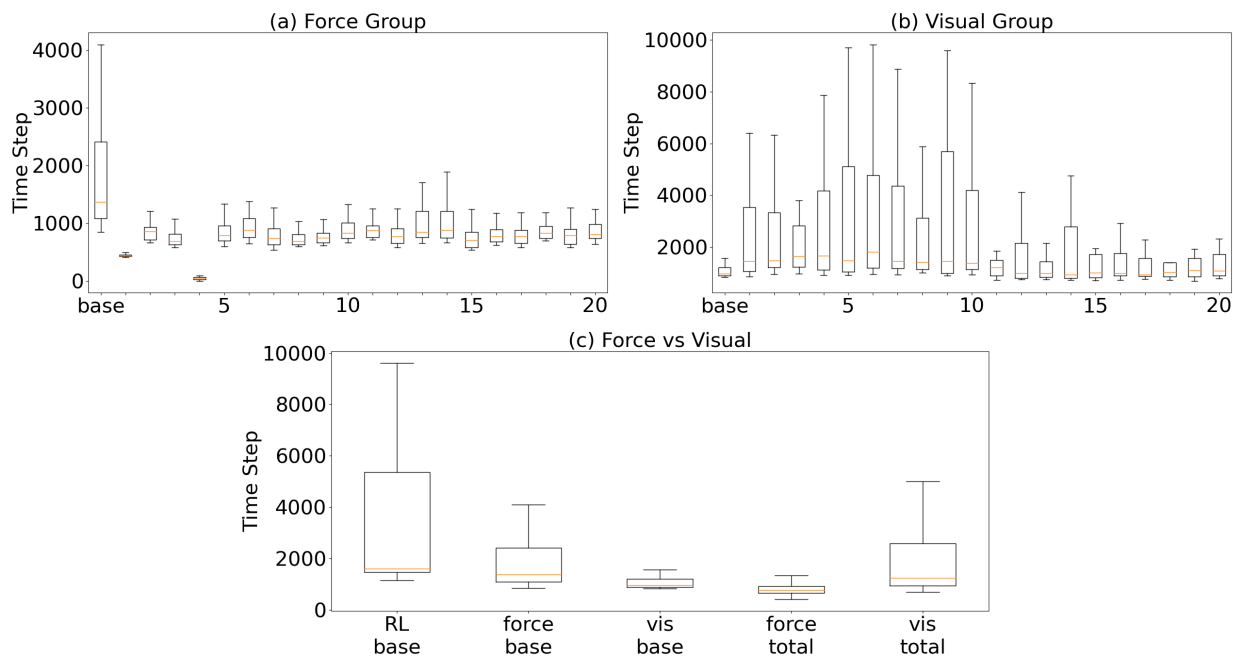

**Fig. 8.** Comparison of distributions for force and visual group inferencing results with inner pipe randomization.

**Table 2.** Test results of inner pipe randomization for force group (100 trials each).

| Trial | RL | Base | 1 | 2 | 3 | 4 | 5 | 6 | 7 | 8 | 9 | 10 |
|---|---|---|---|---|---|---|---|---|---|---|---|---|
| Success | 24 | 40 | 100 | 80 | 94 | 89 | 75 | 79 | 85 | 41 | 83 | 80 |
| Mean | 3500 | 2149 | 458 | 918 | 973 | 46 | 1015 | 994 | 783 | 2119 | 848 | 794 |
| STD | 55.20 | 40.23 | 9.83 | 21.45 | 34.06 | 5.04 | 32.96 | 19.95 | 14.08 | 95.85 | 23.78 | 12.43 |
| Trial | 11 | 12 | 13 | 14 | 15 | 16 | 17 | 18 | 19 | 20 | | |
| Success | 77` | 95 | 85 | 76 | 88 | 79 | 90 | 84 | 92 | 39 | | |
| Mean | 1046 | 936 | 1930 | 2464 | 788 | 1411 | 894 | 941 | 910 | 2026 | | |
| STD | 29.90 | 28.49 | 55.02 | 89.26 | 19.41 | 41.93 | 29.89 | 20.22 | 30.16 | 79.10 | | |

**Table 4**. Test results of inner pipe randomization for visual group (100 trials each).

| Trial | RL | Base | 1 | 2 | 3 | 4 | 5 | 6 | 7 | 8 | 9 | 10 |
|---|---|---|---|---|---|---|---|---|---|---|---|---|
| Success | 24 | 15 | 32 | 24 | 20 | 27 | 21 | 19 | 16 | 43 | 30 | 32 |
| Mean | 3500 | 1812 | 2366 | 3024 | 2613 | 2796 | 2952 | 3313 | 2794 | 2659 | 3304 | 2952 |
| STD | 55.20 | 47.01 | 43.11 | 53.74 | 49.28 | 49.27 | 51.68 | 52.48 | 48.73 | 46.22 | 54.19 | 50.93 |
| Trial | 11 | 12 | 13 | 14 | 15 | 16 | 17 | 18 | 19 | 20 | | |
| Success | 28 | 27 | 22 | 29 | 21 | 32 | 27 | 24 | 35 | 42 | | |
| Mean | 1461 | 1529 | 1494 | 1740 | 1665 | 1456 | 1564 | 1513 | 1515 | 1544 | | |
| STD | 29.78 | 31.95 | 34.88 | 36.68 | 36.89 | 29.68 | 35.41 | 33.17 | 33.97 | 32.28 | | |

**Fig.8** showed the box plots of test results for the force group and visual group. The length of successful trials for force trials followed a more concentrated distribution compared with visual groups, indicating that the force demonstrations, even with random initialization of pipe positions, could still give stable and effective instruction to the training process. In other words, the policy guided by force demonstration is more likely to be the optimal solution given a certain training scenario compared with the visual instruction.

*5.3. Condition 2: Randomized Target Location*

Then, we further randomized the outer pipe and tested all the trials following the workflow discussed above. The distance range between the target center and the initial position is from 0.4 m to 1.0 m along the x-axis and within a 0.2 m radius on the y-z plane. we randomized the outer pipe and the target location and tested all the trials following the workflow discussed above. Similar to the previous experiment, we conducted 100 inferences for each trial and the results were listed in **Table 4** and **Table 5**. **Fig.9** further shows the distributions of force and visual group. Since both the inner and outer pipe positions were randomly initialized, the successful rates were lower than those in **Table 2** and **Table 3**, but the force trials' results were still better than that of visual trials. Additionally, the average success rate was 74.1% of the force group and 25.7% of the visual group. Compared to 81.2% and 27.6% from the previous experiment, the successful rate for the force group slightly dropped but was still at a relatively high level, which verified the generality of force demonstration. On the contrary, the successful rates for the visual group were all at a low level, which further proved that the visual instructions were not helpful in guiding small-scale contact-rich dexterous operation.

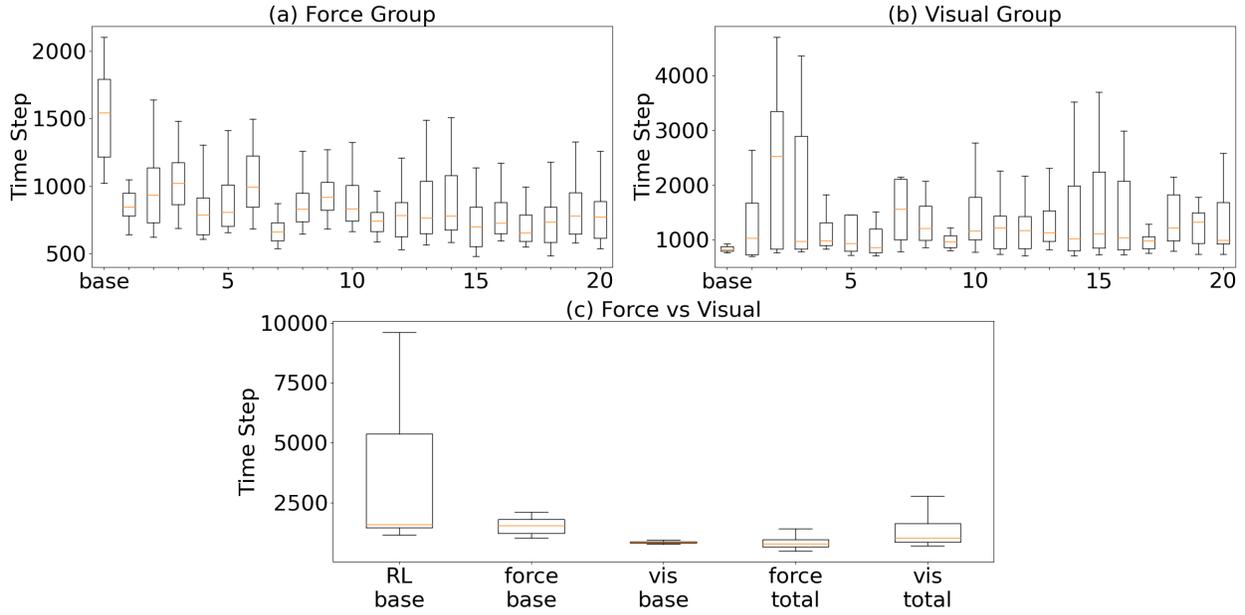

**Fig. 9.** Comparison of distributions for force and visual group inferencing results with outer pipe randomization.

Table 4. Test results of target randomization for force group (100 trials each).

| Trial | RL | Base | 1 | 2 | 3 | 4 | 5 | 6 | 7 | 8 | 9 | 10 |
|---|---|---|---|---|---|---|---|---|---|---|---|---|
| **Success** | 14 | 46 | 74 | 76 | 76 | 88 | 80 | 70 | 84 | 35 | 72 | 72 |
| **Mean** | 2141 | 1882 | 905 | 995 | 1063 | 848 | 871 | 1140 | 699 | 1914 | 945 | 932 |
| **STD** | 32.98 | 33.49 | 14.98 | 19.62 | 17.12 | 16.34 | 14.25 | 23.02 | 11.96 | 49.73 | 13.84 | 17.02 |
| **Trial** | **11** | **12** | **13** | **14** | **15** | **16** | **17** | **18** | **19** | **20** | | |
| **Success** | 67 | 98 | 72 | 73 | 86 | 77 | 76 | 94 | 76 | 31 | | |
| **Mean** | 897 | 948 | 1082 | 1144 | 756 | 902 | 1011 | 763 | 1008 | 1671 | | |
| **STD** | 27.70 | 31.23 | 29.76 | 31.39 | 21.38 | 22.20 | 32.78 | 15.63 | 33.67 | 67.31 | | |

Table 5. Test results of target randomization for the visual group (100 trials each).

| Trial | RL | Base | 1 | 2 | 3 | 4 | 5 | 6 | 7 | 8 | 9 | 10 |
|---|---|---|---|---|---|---|---|---|---|---|---|---|
| **Success** | 14 | 6 | 16 | 26 | 18 | 34 | 28 | 24 | 20 | 38 | 16 | 32 |
| **Mean** | 2141 | 842 | 1479 | 2218 | 1825 | 1208 | 1467 | 1152 | 1698 | 1621 | 1032 | 1645 |
| STD | 32.98 | 8.21 | 31.63 | 37.04 | 36.75 | 21.44 | 32.82 | 26.17 | 29.74 | 32.11 | 15.57 | 31.68 |
| **Trial** | **11** | **12** | **13** | **14** | **15** | **16** | **17** | **18** | **19** | **20** | | |
| **Success** | 24 | 26 | 24 | 29 | 28 | 19 | 32 | 24 | 28 | 33 | | |
| **Mean** | 1469 | 1397 | 1428 | 1626 | 1707 | 1655 | 1059 | 1605 | 1451 | 1490 | | |
| **STD** | 29.26 | 29.51 | 27.83 | 34.86 | 33.85 | 34.86 | 20.43 | 31.75 | 26.13 | 29.88 | | |

*5.4. Force Group Performance Variation*

Our investigation into force-based imitation learning has revealed significant advantages over visual-based methods, particularly in terms of training efficacy and accuracy of outcomes. Nonetheless, a closer inspection of individual results indicates that the benefits of force feedback are not uniform across all demonstrations. Specifically, demonstration 8 and demonstration 20 from force group exhibited slower convergence during training compared with the corresponding ones from visual groups, as depicted by the blue and red curves in **Fig.10**. Unlike the training results from the others, the agents trained by these two force demonstrations has worse performance compared with the agents trained by the corresponding visual groups (8 and 20). Although both the two force demonstrations lead the agents to converge at the same level compared with the visual demonstrations, the converging speeds for force demonstrations at the beginning phase (0-50) epochs were much slower than the corresponding visual training curves. Additionally, their inference success rates are slightly lower than the results from their corresponding visual demonstrations. The success rate is 41% for force demonstration 8 and 39% for force demonstration 20, whereas the success rates for the corresponding visual demonstrations are 43% and 42%, respectively. For condition two, the success rates for the two forces demonstrations are 35% and 31%, which are also lower than the visual demonstrations with 38% and 33%. Furthermore, the mean values of complete steps and STDs of these two demonstrations within the 100 inference trials are larger than the rest of force group, indicating that the models guided by these two demonstrations tend to have unstable behaviors. The results are indicated by the red markings in **Tables 2-5**.

This variation within the force group suggests that while force feedback can generally enhance performance, it is also susceptible to fluctuations that can drastically affect the success of the learning process. Force demonstrations can result in a broader spectrum of performance outcomes. The insight pointed to the importance to understand and address the variation in force demonstration performance. Identifying the factors that contribute to this variability will be crucial for developing more reliable and consistent training methods for force-based imitation learning.

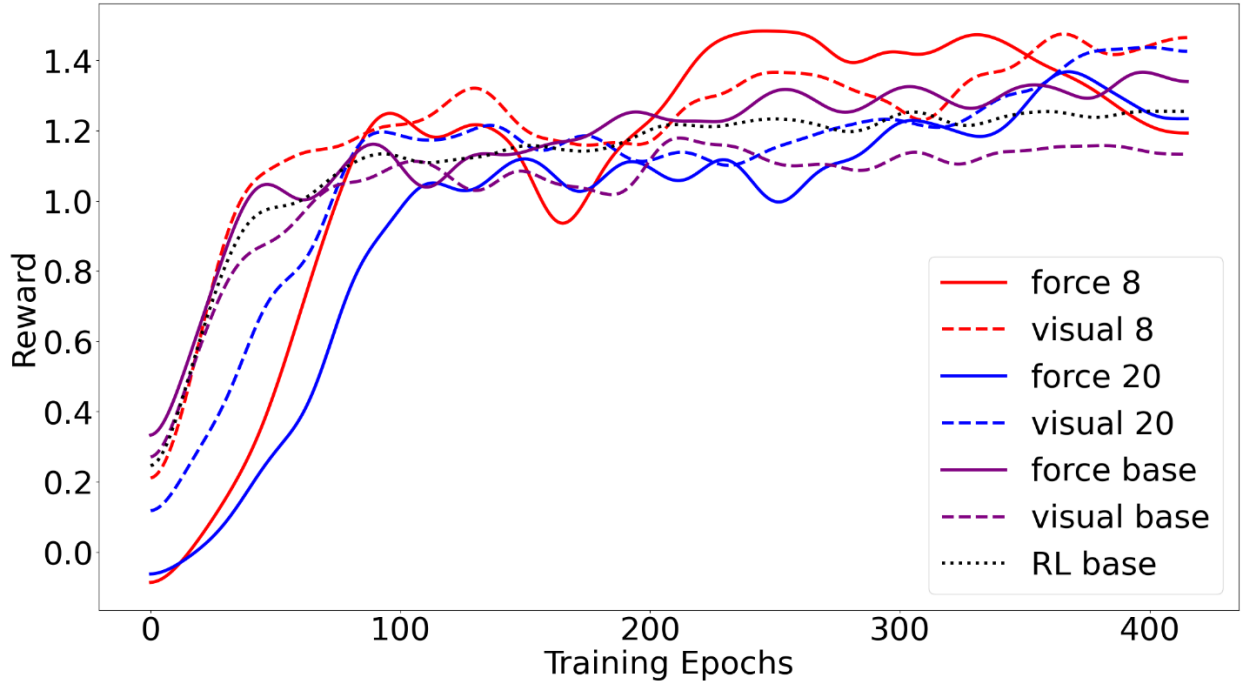

**Fig. 10.** Highlights of Force trial 8 and trail 20.

## 6. DISCUSSION AND CONCLUSIONS

The momentum for deploying automation in the construction industry has emphasized the potential of robots in various construction activities. However, tasks like welding and pipefitting that require dexterous manipulations present significant challenges due to the need for delicate handling and adaptive force controls. These intricacies make training robots particularly complex, especially when traditional methods rely primarily on vision-based or position-based demonstrations. Our study introduces a novel two-phase system framework that integrates force feedback data from human operators with robotic learning processes, addressing the limitations of visual and positional cues.

The core innovation of our framework lies in the data-bridge module and the use of Generative Adversarial Imitation Learning (GAIL), uniquely incorporating force feedback into the robot's reward function. This approach significantly enhances training, particularly in scenarios where visual and positional cues are limited. Our results show that robots trained with force feedback achieve higher accuracy and better task completion rates than those relying solely on visual cues.

Force feedback captures complex human motor strategies, making it crucial for high-dexterity tasks. This study demonstrates the superiority of force feedback over visual methods, with robots trained in this way showing greater adaptability and generalization. These findings have transformative potential for industries like construction, where precision and adaptability are key to operational success.

However, it's important to acknowledge the limitations inherent in imitation learning, notably its heavy reliance on high-quality demonstration data. Incomplete or unclear demonstrations can result in suboptimal learning, and the subjective nature of human demonstrations can introduce biases that affect generalizability. Additionally, while our optimized force simulation system reduces computational load and enhances real-time performance in contact-rich tasks, it may oversimplify complex dynamic interactions, such as detailed force-deformation behaviors. The custom mathematical model used for force feedback in the pipe insertion task is efficient but may struggle in more diverse scenarios, limiting versatility in broader applications.

To address these challenges, our future work will focus on reducing biases in demonstration data, refining the system to generalize across tasks and operators. We will incorporate more advanced collision models that better capture detailed force-deformation interactions and improve the system's dynamic response to reflect real-world conditions while maintaining real-time performance. Additionally, we plan to validate the learning framework with real-world data by designing sensor systems to capture accurate force feedback and optimize the transfer from simulation to real-world robot control.

In conclusion, our study advocates for the integration of authentic force feedback in reinforcement learning for tasks that demand high dexterity and precision. The demonstrated advantages of force feedback over visual feedback in terms of accuracy, success rate, and adaptability highlight its potential as a pivotal element in future human-robot interaction paradigms. Bridging the gap between virtual training and real-world deployment remains an exciting future research avenue in the intersection of embodied AI and robotics.

ACKNOWLEDGEMENTS

This material is supported by the National Science Foundation (NSF) under grant 2024784. Any opinions, findings, conclusions, or recommendations expressed in this article are those of the authors and do not reflect the views of the NSF.

**REFERENCES**

[1] S.O. Abioye, L.O. Oyedele, L. Akanbi, A. Ajayi, J.M.D. Delgado, M. Bilal, O.O. Akinade, A. Ahmed, Artificial intelligence in the construction industry: A review of present status, opportunities and future challenges, Journal of Building Engineering, 44 (2021) 103299.

[2] S.K. Baduge, S. Thilakarathna, J.S. Perera, M. Arashpour, P. Sharafi, B. Teodosio, A. Shringi, P. Mendis, Artificial intelligence and smart vision for building and construction 4.0: Machine and deep learning methods and applications, Automation in Construction, 141 (2022) 104440.

[3] P. Pradhananga, M. ElZomor, G. Santi Kasabdji, Identifying the challenges to adopting robotics in the US construction industry, Journal of Construction Engineering and Management, 147 (2021) 05021003.

[4] S. Halder, K. Afsari, Robots in inspection and monitoring of buildings and infrastructure: A systematic review, Applied Sciences, 13 (2023) 2304.

[5] S. Sudhakaran, K. Montgomery, M. Kashef, D. Cavalcanti, R. Candell, Wireless time sensitive networking for industrial collaborative robotic workcells, 2021 17th IEEE International Conference on Factory Communication Systems (WFCS), IEEE, 2021, pp. 91-94.

[6] S. Surati, S. Hedaoo, T. Rotti, V. Ahuja, N. Patel, Pick and place robotic arm: a review paper, Int. Res. J. Eng. Technol, 8 (2021) 2121-2129.

[7] U.-H. Kim, J.-M. Park, T.-J. Song, J.-H. Kim, 3-D scene graph: A sparse and semantic representation of physical environments for intelligent agents, IEEE transactions on cybernetics, 50 (2019) 4921-4933.

[8] M. Matulis, C. Harvey, A robot arm digital twin utilising reinforcement learning, Computers & Graphics, 95 (2021) 106-114.

[9] Y. Ye, T. Zhou, J. Du, Robot-assisted immersive kinematic experience transfer for welding training, Journal of Computing in Civil Engineering, 37 (2023) 04023002.

[10] T. Zhou, Embodied Robot Teleoperation in Construction, University of Florida, 2023.

[11] Y. Hengxu, Z. Tianyu, Z. Qi, Y. Yang, D. Eric Jing, Embodied AI for Dexterity Capable Construction Robots: DEXBOT Framework, 2023.

[12] Z.-H. Yin, W. Ye, Q. Chen, Y. Gao, Planning for Sample Efficient Imitation Learning, Advances in Neural Information Processing Systems, 35 (2022) 2577-2589.

[13] A. Hussein, M.M. Gaber, E. Elyan, C. Jayne, Imitation learning: A survey of learning methods, ACM Computing Surveys (CSUR), 50 (2017) 1-35.

[14] H. Yu, V.R. Kamat, C.C. Menassa, Cloud-Based Hierarchical Imitation Learning for Scalable Transfer of Construction Skills from Human Workers to Assisting Robots, Journal of Computing in Civil Engineering, 38 (2024) 04024019.

[15] J. Ho, S. Ermon, Generative adversarial imitation learning, Advances in neural information processing

systems, 29 (2016).

[16] F. Torabi, G. Warnell, P. Stone, Recent advances in imitation learning from observation, arXiv preprint arXiv:1905.13566, (2019).

[17] C. Finn, T. Yu, T. Zhang, P. Abbeel, S. Levine, One-shot visual imitation learning via meta-learning, Conference on robot learning, PMLR, 2017, pp. 357-368.

[18] Z. Xia, Z. Deng, B. Fang, Y. Yang, F. Sun, A review on sensory perception for dexterous robotic manipulation, International Journal of Advanced Robotic Systems, 19 (2022) 17298806221095974.

[19] Q. Zhu, T. Zhou, J. Du, Haptics-based force balance controller for tower crane payload sway controls, Automation in Construction, 144 (2022) 104597.

[20] A.A. Valdivia, R. Shailly, N. Seth, F. Fuentes, D.P. Losey, L.H. Blumenschein, Wrapped haptic display for communicating physical robot learning, 2022 IEEE 5th International Conference on Soft Robotics (RoboSoft), IEEE, 2022, pp. 823-830.

[21] I. El Rassi, J.-M. El Rassi, A review of haptic feedback in tele-operated robotic surgery, Journal of medical engineering & technology, 44 (2020) 247-254.

[22] L. Chen, H. Luo, Near-optimal goal-oriented reinforcement learning in non-stationary environments, Advances in Neural Information Processing Systems, 35 (2022) 33973-33984.

[23] B. Singh, R. Kumar, V.P. Singh, Reinforcement learning in robotic applications: a comprehensive survey, Artificial Intelligence Review, (2022) 1-46.

[24] B. Zheng, S. Verma, J. Zhou, I. Tsang, F. Chen, Imitation learning: Progress, taxonomies and challenges, arXiv preprint arXiv:2106.12177, (2021).

[25] K. Arulkumaran, D.O. Lillrank, A pragmatic look at deep imitation learning, arXiv preprint arXiv:2108.01867, (2021).

[26] M. Liu, W. Buntine, G. Haffari, Learning how to actively learn: A deep imitation learning approach, Proceedings of the 56th Annual Meeting of the Association for Computational Linguistics (Volume 1: Long Papers), 2018, pp. 1874-1883.

[27] D. Lee, S. Lee, N. Masoud, M. Krishnan, V.C. Li, Digital twin-driven deep reinforcement learning for adaptive task allocation in robotic construction, Advanced Engineering Informatics, 53 (2022) 101710.

[28] K. Xu, Z. Hu, R. Doshi, A. Rovinsky, V. Kumar, A. Gupta, S. Levine, Dexterous manipulation from images: Autonomous real-world rl via substep guidance, 2023 IEEE International Conference on Robotics and Automation (ICRA), IEEE, 2023, pp. 5938-5945.

[29] G.B. Margolis, G. Yang, K. Paigwar, T. Chen, P. Agrawal, Rapid Locomotion via Reinforcement Learning.

[30] J. Hua, L. Zeng, G. Li, Z. Ju, Learning for a robot: Deep reinforcement learning, imitation learning, transfer learning, Sensors, 21 (2021) 1278.

[31] J. Ibarz, J. Tan, C. Finn, M. Kalakrishnan, P. Pastor, S. Levine, How to train your robot with deep reinforcement learning: lessons we have learned, The International Journal of Robotics Research, 40 (2021) 698-721.

[32] J. Tarbouriech, M. Pirotta, M. Valko, A. Lazaric, A provably efficient sample collection strategy for reinforcement learning, Advances in Neural Information Processing Systems, 34 (2021) 7611-7624.

[33] P. Wenzel, T. Schön, L. Leal-Taixé, D. Cremers, Vision-based mobile robotics obstacle avoidance with deep reinforcement learning, 2021 IEEE International Conference on Robotics and Automation (ICRA),

IEEE, 2021, pp. 14360-14366.
[34] M. Everett, Y.F. Chen, J.P. How, Collision avoidance in pedestrian-rich environments with deep reinforcement learning, IEEE Access, 9 (2021) 10357-10377.
[35] G. Kahn, A. Villaflor, B. Ding, P. Abbeel, S. Levine, Self-supervised deep reinforcement learning with generalized computation graphs for robot navigation, 2018 IEEE international conference on robotics and automation (ICRA), IEEE, 2018, pp. 5129-5136.
[36] C. Pérez-D'Arpino, C. Liu, P. Goebel, R. Martín-Martín, S. Savarese, Robot navigation in constrained pedestrian environments using reinforcement learning, 2021 IEEE International Conference on Robotics and Automation (ICRA), IEEE, 2021, pp. 1140-1146.
[37] T. Zhou, Q. Zhu, J. Du, Intuitive robot teleoperation for civil engineering operations with virtual reality and deep learning scene reconstruction, Advanced Engineering Informatics, 46 (2020) 101170.
[38] D. Liu, Z. Wang, B. Lu, M. Cong, H. Yu, Q. Zou, A reinforcement learning-based framework for robot manipulation skill acquisition, IEEE Access, 8 (2020) 108429-108437.
[39] A. El-Fakdi, M. Carreras, Two-step gradient-based reinforcement learning for underwater robotics behavior learning, Robotics and Autonomous Systems, 61 (2013) 271-282.
[40] P. Xia, F. Xu, T. Zhou, J. Du, Benchmarking human versus robot performance in emergency structural inspection, Journal of Construction Engineering and Management, 148 (2022) 04022070.
[41] F. Niroui, K. Zhang, Z. Kashino, G. Nejat, Deep reinforcement learning robot for search and rescue applications: Exploration in unknown cluttered environments, IEEE Robotics and Automation Letters, 4 (2019) 610-617.
[42] J. Hao, T. Yang, H. Tang, C. Bai, J. Liu, Z. Meng, P. Liu, Z. Wang, Exploration in deep reinforcement learning: From single-agent to multiagent domain, IEEE Transactions on Neural Networks and Learning Systems, (2023).
[43] J.M.D. Delgado, L. Oyedele, Robotics in construction: A critical review of the reinforcement learning and imitation learning paradigms, Advanced Engineering Informatics, 54 (2022) 101787.
[44] D.K. Jha, S. Jain, D. Romeres, W. Yerazunis, D. Nikovski, Generalizable human-robot collaborative assembly using imitation learning and force control, 2023 European Control Conference (ECC), IEEE, 2023, pp. 1-8.
[45] A. Sasagawa, K. Fujimoto, S. Sakaino, T. Tsuji, Imitation learning based on bilateral control for human–robot cooperation, IEEE Robotics and Automation Letters, 5 (2020) 6169-6176.
[46] C.-J. Liang, V.R. Kamat, C.C. Menassa, Teaching robots to perform quasi-repetitive construction tasks through human demonstration, Automation in Construction, 120 (2020) 103370.
[47] Y. Lu, J. Fu, G. Tucker, X. Pan, E. Bronstein, B. Roelofs, B. Sapp, B. White, A. Faust, S. Whiteson, Imitation is not enough: Robustifying imitation with reinforcement learning for challenging driving scenarios, arXiv preprint arXiv:2212.11419, (2022).
[48] S. Fujimoto, S.S. Gu, A minimalist approach to offline reinforcement learning, Advances in neural information processing systems, 34 (2021) 20132-20145.
[49] T. Hester, M. Vecerik, O. Pietquin, M. Lanctot, T. Schaul, B. Piot, D. Horgan, J. Quan, A. Sendonaris, I. Osband, Deep q-learning from demonstrations, Proceedings of the AAAI conference on artificial intelligence, 2018.
[50] A. Rajeswaran, V. Kumar, A. Gupta, G. Vezzani, J. Schulman, E. Todorov, S. Levine, Learning complex

dexterous manipulation with deep reinforcement learning and demonstrations, arXiv preprint arXiv:1709.10087, (2017).
[51] M. Vecerik, T. Hester, J. Scholz, F. Wang, O. Pietquin, B. Piot, N. Heess, T. Rothörl, T. Lampe, M. Riedmiller, Leveraging demonstrations for deep reinforcement learning on robotics problems with sparse rewards, arXiv preprint arXiv:1707.08817, (2017).
[52] L. Huang, Z. Zhu, Z. Zou, To imitate or not to imitate: Boosting reinforcement learning-based construction robotic control for long-horizon tasks using virtual demonstrations, Automation in Construction, 146 (2023) 104691.
[53] K.M. Lundeen, V.R. Kamat, C.C. Menassa, W. McGee, Scene understanding for adaptive manipulation in robotized construction work, Automation in Construction, 82 (2017) 16-30.
[54] K.M. Lundeen, V.R. Kamat, C.C. Menassa, W. McGee, Autonomous motion planning and task execution in geometrically adaptive robotized construction work, Automation in Construction, 100 (2019) 24-45.
[55] X. Wang, C.-J. Liang, C.C. Menassa, V.R. Kamat, Interactive and immersive process-level digital twin for collaborative human–robot construction work, Journal of Computing in Civil Engineering, 35 (2021) 04021023.
[56] C.-J. Liang, W. McGee, C. Menassa, V. Kamat, Bi-directional communication bridge for state synchronization between digital twin simulations and physical construction robots, Proceedings of the International Symposium on Automation and Robotics in Construction (IAARC), 2020.
[57] C.-J. Liang, X. Wang, V.R. Kamat, C.C. Menassa, Human–robot collaboration in construction: Classification and research trends, Journal of Construction Engineering and Management, 147 (2021) 03121006.
[58] J. Cai, A. Du, X. Liang, S. Li, Prediction-based path planning for safe and efficient human–robot collaboration in construction via deep reinforcement learning, Journal of Computing in Civil Engineering, 37 (2023) 04022046.
[59] C.-J. Liang, K.M. Lundeen, W. McGee, C.C. Menassa, S. Lee, V.R. Kamat, A vision-based marker-less pose estimation system for articulated construction robots, Automation in Construction, 104 (2019) 80-94.
[60] C.-J. Liang, S.-C. Kang, M.-H. Lee, RAS: a robotic assembly system for steel structure erection and assembly, International Journal of Intelligent Robotics and Applications, 1 (2017) 459-476.
[61] C.-J. Liang, V.R. Kamat, C.C. Menassa, W. McGee, Trajectory-based skill learning for overhead construction robots using generalized cylinders with orientation, Journal of Computing in Civil Engineering, 36 (2022) 04021036.
[62] C.-J. Liang, V. Kamat, C. Menassa, Teaching robots to perform construction tasks via learning from demonstration, Proceedings of the 36th International Symposium on Automation and Robotics in Construction, ISARC, 2019, pp. 1305-1311.
[63] K. Asadi, V.R. Haritsa, K. Han, J.-P. Ore, Automated object manipulation using vision-based mobile robotic system for construction applications, Journal of Computing in Civil Engineering, 35 (2021) 04020058.
[64] B. Argall, B. Browning, M. Veloso, Learning by demonstration with critique from a human teacher, Proceedings of the ACM/IEEE international conference on Human-robot interaction, 2007, pp. 57-64.

[65] H. Karnan, G. Warnell, X. Xiao, P. Stone, Voila: Visual-observation-only imitation learning for autonomous navigation, 2022 International Conference on Robotics and Automation (ICRA), IEEE, 2022, pp. 2497-2503.

[66] H. Kim, Y. Ohmura, A. Nagakubo, Y. Kuniyoshi, Training robots without robots: deep imitation learning for master-to-robot policy transfer, IEEE Robotics and Automation Letters, 8 (2023) 2906-2913.

[67] K. Gandhi, S. Karamcheti, M. Liao, D. Sadigh, Eliciting compatible demonstrations for multi-human imitation learning, Conference on Robot Learning, PMLR, 2023, pp. 1981-1991.

[68] O. Mees, L. Hermann, W. Burgard, What matters in language conditioned robotic imitation learning over unstructured data, IEEE Robotics and Automation Letters, 7 (2022) 11205-11212.

[69] B. Fang, S. Jia, D. Guo, M. Xu, S. Wen, F. Sun, Survey of imitation learning for robotic manipulation, International Journal of Intelligent Robotics and Applications, 3 (2019) 362-369.

[70] H. Yin, A. Varava, D. Kragic, Modeling, learning, perception, and control methods for deformable object manipulation, Science Robotics, 6 (2021) eabd8803.

[71] Q. Wu, X. Gong, K. Xu, D. Manocha, J. Dong, J. Wang, Towards target-driven visual navigation in indoor scenes via generative imitation learning, IEEE Robotics and Automation Letters, 6 (2020) 175-182.

[72] M. Doering, D.F. Glas, H. Ishiguro, Modeling interaction structure for robot imitation learning of human social behavior, IEEE Transactions on Human-Machine Systems, 49 (2019) 219-231.

[73] R. Li, Z. Zou, Enhancing construction robot learning for collaborative and long-horizon tasks using generative adversarial imitation learning, Advanced Engineering Informatics, 58 (2023) 102140.

[74] M. Kappel, V. Golyanik, M. Elgharib, J.-O. Henningson, H.-P. Seidel, S. Castillo, C. Theobalt, M. Magnor, High-fidelity neural human motion transfer from monocular video, Proceedings of the IEEE/CVF conference on computer vision and pattern recognition, 2021, pp. 1541-1550.

[75] S. Li, H. Yu, W. Ding, H. Liu, L. Ye, C. Xia, X. Wang, X.-P. Zhang, Visual–Tactile Fusion for Transparent Object Grasping in Complex Backgrounds, IEEE Transactions on Robotics, (2023).

[76] T. Anzai, K. Takahashi, Deep gated multi-modal learning: In-hand object pose changes estimation using tactile and image data, 2020 IEEE/RSJ International Conference on Intelligent Robots and Systems (IROS), IEEE, 2020, pp. 9361-9368.

[77] T. Boroushaki, J. Leng, I. Clester, A. Rodriguez, F. Adib, Robotic grasping of fully-occluded objects using rf perception, 2021 IEEE International Conference on Robotics and Automation (ICRA), IEEE, 2021, pp. 923-929.

[78] C. Yu, P. Wang, Dexterous manipulation for multi-fingered robotic hands with reinforcement learning: a review, Frontiers in Neurorobotics, 16 (2022) 861825.

[79] H. Yu, V.R. Kamat, C.C. Menassa, W. McGee, Y. Guo, H. Lee, Mutual physical state-aware object handover in full-contact collaborative human-robot construction work, Automation in Construction, 150 (2023) 104829.

[80] M. Lambeta, P.-W. Chou, S. Tian, B. Yang, B. Maloon, V.R. Most, D. Stroud, R. Santos, A. Byagowi, G. Kammerer, Digit: A novel design for a low-cost compact high-resolution tactile sensor with application to in-hand manipulation, IEEE Robotics and Automation Letters, 5 (2020) 3838-3845.